\documentclass[10pt,twocolumn,letterpaper]{article}

\usepackage[pagenumbers]{cvpr} 
 \usepackage{multirow}
\usepackage[dvipsnames]{xcolor}

\definecolor{cvprblue}{rgb}{0.21,0.49,0.74}
\usepackage[pagebackref,breaklinks,colorlinks,citecolor=cvprblue]{hyperref}

\def\paperID{*****} 
\def\confName{CVPR}
\def\confYear{2024}

\title{Detecting Generative Parroting through Overfitting Masked Autoencoders}

\author{Saeid Asgari Taghanaki\\
Autodesk AI Research\\
{\tt\small saeid.asgari.taghanaki@autodesk.com}
\and
Joseph Lambourne\\
Autodesk AI Research\\
{\tt\small joseph.lambourne@autodesk.com}
}

\begin{document}
\maketitle
\begin{abstract}
The advent of generative AI models has revolutionized digital content creation, yet it introduces challenges in maintaining copyright integrity due to generative parroting, where models mimic their training data too closely. Our research presents a novel approach to tackle this issue by employing an overfitted Masked Autoencoder (MAE) to detect such parroted samples effectively. We establish a detection threshold based on the mean loss across the training dataset, allowing for the precise identification of parroted content in modified datasets. Preliminary evaluations demonstrate promising results, suggesting our method's potential to ensure ethical use and enhance the legal compliance of generative models.

\end{abstract}    
\section{Introduction}
\label{sec:intro}

Generative artificial intelligence (AI) models, including but not limited to Stable Diffusion~\cite{rombach2022high}, DALLE~\cite{ramesh2021zero}, and Generative Pre-trained Transformers (GPT)~\cite{mann2020language}, represent a groundbreaking shift in the landscape of digital content creation, empowering users to generate text, images, and other forms of media with unprecedented ease and flexibility. These models have been applied across a wide range of domains, from artistic creation and design to content generation for social media and marketing purposes, demonstrating their versatility and potential to enhance creativity and productivity.

The rapid adoption and deployment of these technologies have also raised significant ethical, legal, and technical challenges, particularly in the context of copyright infringement and data privacy~\cite{franceschelli2022copyright,sag2023copyright,vyas2023provable}. At the heart of these concerns is the phenomenon known as ``generative parroting," where models produce outputs that are not sufficiently distinct from their training data~\cite{carlini2023extracting,somepalli2023diffusion}, leading to the generation of content that closely mimics or even directly copies existing copyrighted materials. This issue not only poses legal risks for users and developers but also undermines trust in generative AI technologies, especially in trust-critical scenarios where the protection of intellectual property and sensitive information is paramount.

The challenge of detecting and mitigating generative parroting is compounded by the inherent complexities of AI models' training processes and the vastness of the data landscapes they navigate. Traditional approaches to model training and evaluation may not adequately address the nuances of copyright-sensitive scenarios, necessitating innovative solutions that are specifically tailored to recognize and respect the boundaries of copyright law~\cite{samuelson2023generative}. Moreover, the dynamic nature of copyright legislation, which varies across jurisdictions and is continually evolving in response to technological advancements, adds another layer of complexity to this challenge~\cite{lucchi2023chatgpt}.

While passing generated samples through a representation learner to obtain feature vectors and compare them with training data might be feasible for small datasets, this approach becomes impractical for larger datasets with billions of samples, especially in real-time scenarios. For instance, designers interacting with generative models need immediate feedback, rendering exhaustive comparisons untenably slow.

In pursuit of an efficient solution, our work proposes the use of a single model capable of encapsulating the essence of the training data and providing a binary response—indicating whether a sample is parroted or not—without the need for pairwise comparison with the entire training dataset. This paper demonstrates that by exploiting the tendency of a Masked Autoencoder (MAE) to overfit, we can effectively identify parroted samples. We hypothesize and confirm that an overfitted MAE discerns between samples that are closely aligned with the training data and those that are novel or substantially altered. The resulting loss value from the reconstruction process acts as an effective metric to distinguish potential instances of parroting. This method offers a significant step towards efficient real-time detection of generative parroting, streamlining the design process and safeguarding the creative output.

By providing a mechanism to detect and flag potential instances of generative parroting, we aim to contribute to the ongoing discourse on ethical AI development and deployment, fostering an environment where generative models can be used responsibly and creatively without compromising copyright integrity or customer trust.
\section{Related Work}

The evaluation and mitigation of generative parroting have been explored in various capacities within the machine learning community. However, these explorations often fall short of providing scalable solutions for detecting parroted content within extensive datasets.

Gulrajani et al.~\cite{gulrajani2020towards} propose benchmarks aimed at resisting trivial memorization by generative models, focusing on the use of neural network divergences for evaluation. While their work contributes valuable insights into model generalization, it does not offer a direct mechanism for identifying individual parroted samples within large datasets.
Vyas, Kakade, and Barak~\cite{vyas2023provable} introduce a formal definition of Near Access-Freeness and present algorithms aimed at copyright protection for generative models by ensuring outputs diverge sufficiently from any potentially copyrighted training data. While this work makes significant theoretical contributions to copyright protection in generative modeling, its practical application in detecting specific instances of parroted content across billions of data points remains computationally challenging.
Meehan, Chaudhuri, and Dasgupta~\cite{meehan2020non} propose a method which signals the presence of data copying across a broad class of models but does not scale effectively to the era of large datasets, as it does not specifically address the computational challenges inherent in analyzing billions of data points.

Carlini et al.'s~\cite{carlini2022quantifying} investigate the extent to which large language models memorize parts of their training data. Their analysis uncovers three log-linear relationships that quantify the degree of emitted memorized training data as a function of the model's capacity, the repetition of examples within the training data, and the amount of context used to prompt the model. They demonstrate that memorization is more prevalent than previously understood and suggest it will likely increase as models continue to scale, highlighting a significant challenge for ensuring privacy and reducing the risk of copyright infringement in generated content. However, their study focuses on language models and the explicit replication of verbatim text, which differs from the broader scope of generative parroting that encompasses various data modalities and subtler forms of content replication addressed by our work with overfitted MAEs. While Carlini et al. provide a foundational understanding of memorization dynamics in neural language models, their approach to quantifying memorization does not directly tackle the computational challenges of detecting parroted content in the vast and diverse datasets typical of today's generative models, underscoring the novelty and necessity of our methodology for scalable detection.

These studies, while instrumental in advancing our understanding of generative model evaluation and the nuances of model memorization, underscore a significant gap: the need for computationally feasible methods to detect parroted content amidst the challenges posed by today's large datasets. The computational overhead of existing approaches, such as feature extraction and comparison across billions of samples, renders them impractical for application in real-world scenarios where dataset sizes can be immense.

Our work seeks to bridge this gap by introducing an overfitted MAE approach, specifically tailored to identify parroted samples without the need for exhaustive dataset comparisons. This method not only addresses the computational inefficiencies of previous models but also opens new avenues for scalable detection of generative parroting across various data modalities.

\section{Methodology}
\label{sec:method}

\begin{figure*}[ht!]
\centering
\includegraphics[width=0.9\linewidth]{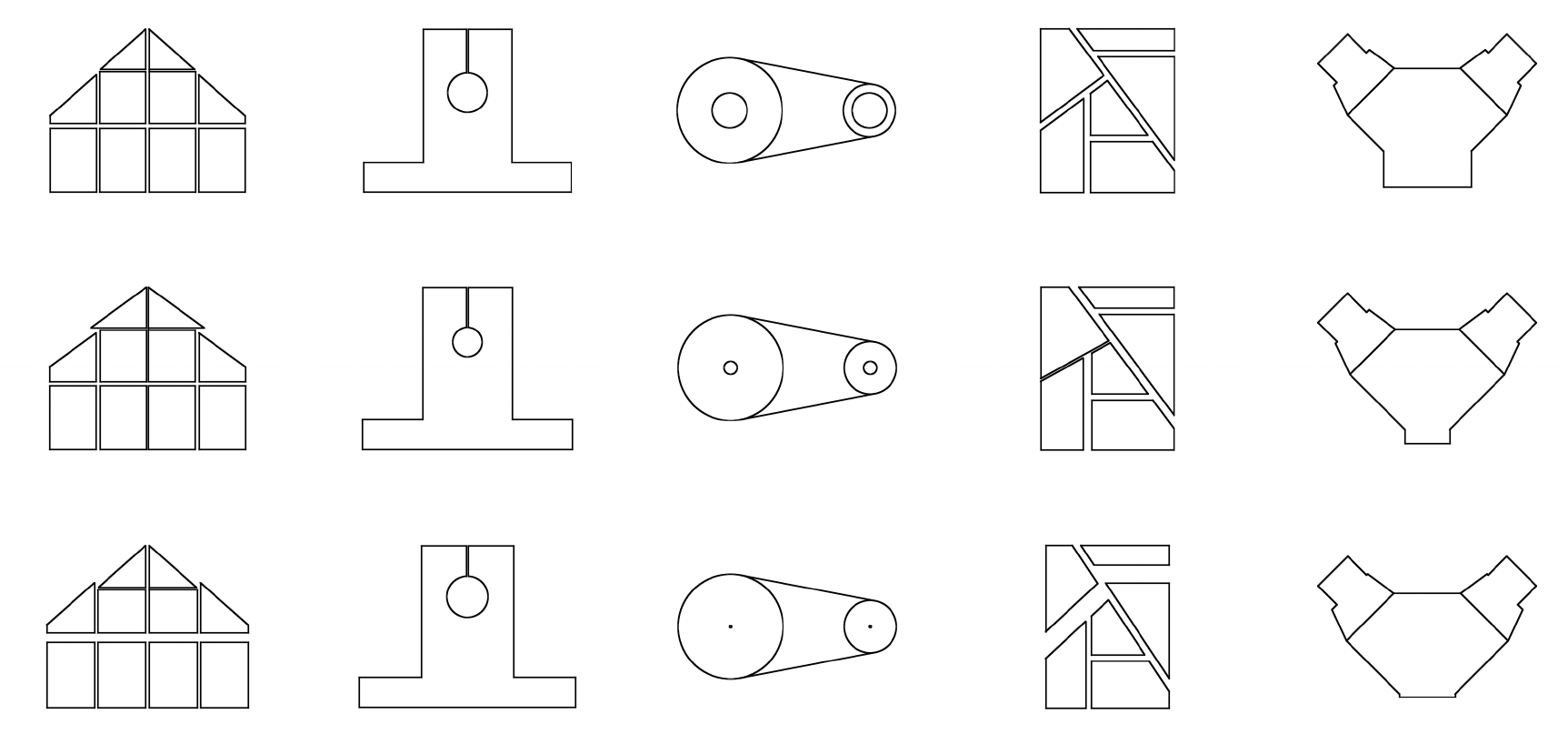}
\caption{Representative samples from the datasets: $D_{\text{train}}$ (original training set), $D_{\text{var 1}}$ (first variation), and $D_{\text{var 2}}$ (second variation), shown from the first to third rows, respectively}
\label{fig:samples}
\end{figure*}

\subsection{Dataset}

For our preliminary experiments, we focus on 2D computer-aided design (CAD) data, employing the SketchGraphs dataset~\cite{seff2020sketchgraphs}. We utilize a total of 535,358 sketches, from which we have created two distinct variations of each original sketch.  The variations were created by adjusting the sketch parameters which control lengths and angles and using the constraint solver in Fusion 360 \cite{fusionsketchapi} to update the geometry while respecting other sketch constraints.  The original and modified geometries were rendered as PNG images of size 640x480 pixels. Our dataset comprises four subsets for training and evaluation, designed to assess the model's ability to detect parroted content and its response to novel samples:

\begin{itemize}
    \renewcommand{\labelitemi}{}
    \item \textbf{Training set} ($D_{\text{train}}$): Consists of the original, unaltered sketches, serving as a baseline for the model's learning process.
    \item \textbf{Modified set 1} ($D_{\text{mod 1}}$): Derived from $D_{\text{train}}$, with each sample slightly modified to emulate the minor variations that a generative model might produce, akin to potential parroted outputs.  The parameters defining lengths were incremented or decremented by $1/20$th of the maximum length of the sketch's bounding box, while parameters defining angles were varied by 1 degree. 
    \item \textbf{Modified set 2} ($D_{\text{mod 2}}$): Further derived from $D_{\text{train}}$, exhibiting more substantial alterations, representing a wider range of potential generative deviations.  The parameters defining lengths were incremented or decremented by $1/5$th of the maximum bounding box length, while parameters defining angles were varied by 4 degree.
    \item \textbf{Novel set} ($D_{\text{nov}}$): Contains completely new samples, unseen by the model during training, to evaluate the model's ability to correctly pass novel samples that are not parroted.
\end{itemize}

The rationale for such dataset structure is predicated on the assumption that a generative model may output exact duplicates or modified versions of the training data. It is imperative that our detection system accurately identifies these instances. The inclusion of $D_{\text{nov}}$ allows us to measure the model's accuracy in discerning novel, non-parroted samples from those that are parroted. The ultimate goal is to ensure the model flags only genuine instances of parroting while permitting novel content, thereby achieving a fine balance between sensitivity and specificity in real-world applications. In figure~\ref{fig:samples}, we have visualized training and modified samples.

\subsection{Masked Autoencoder (MAE) Loss}

The MAE is designed based on a vision transformer architecture, tasked with processing masked versions of the input data $X$ to reconstruct the original inputs. The reconstruction loss for an input image $X$ and its reconstructed version $\hat{X}$ is calculated using the Mean Squared Error (MSE) over the unmasked portions of the image. Specifically, for the CAD dataset where most of the background pixels are white, the MSE is calculated only on the drawings, i.e., non-fully-white pixels:

\begin{equation}
L_{\text{MSE}}(X, \hat{X}) = \frac{1}{N_{\text{drawing}}} \sum_{i \in N_{\text{drawing}}} (X_i - \hat{X}_i)^2
\end{equation}

where $N_{\text{drawing}}$ represents the number of non-fully-white pixels in the image, and the summation runs over these pixels only. For natural images, the regular MSE calculation over all pixels can be applied.




\subsection{Overfitting and Threshold Setting}

To induce overfitting, the MAE is trained on $D_{\text{train}}$ until the loss on this dataset reaches a minimal value. The threshold $\tau$ for detecting parroted samples is set as the mean loss over the Training Dataset:

\begin{equation}
    \tau = \mu_{L_{\text{train}}} = \frac{1}{|D_{\text{train}}|} \sum_{X \in D_{\text{train}}} L(X, \hat{X})
\end{equation}

A sample is flagged as parroted if its reconstruction loss $L(X, \hat{X})$ is less than or equal to $\tau$. This criterion is applied across $D_{\text{mod 1}}$ and $D_{\text{mod 2}}$ to detect parroted samples.

\section{Experiments}

\renewcommand{\arraystretch}{1.2}
\begin{table*}[h!]
\centering
\caption{Detection results across four datasets: $D_{\text{train}}$, $D_{\text{mod 1}}$, $D_{\text{mod 2}}$, and $D_{\text{nov}}$. 'WD' indicates weight decay, set to 0.05 when enabled. 'AUG' denotes data augmentation; when enabled, only vertical and horizontal flips are used. The '$D_{\text{nov}}$ pass' percentage reflects the novel samples not flagged as parroted.}
\begin{tabular}{cccccccccc}
\hline
\multirow{2}{*}{Method} & \multirow{2}{*}{Num bins} &\multirow{2}{*}{WD} & \multirow{2}{*}{AUG} & \multirow{2}{*}{p\_mask (\%)} & \multirow{2}{*}{Epochs} & \multicolumn{3}{c}{Detection rate (\%)} & \multirow{2}{*}{$D_{\text{nov}}$ pass (\%)} \\ \cline{7-9}
                    &                      &                          &      &  &                   & $D_{\text{train}}$    & $D_{\text{mod 1}}$   & $D_{\text{mod 2}}$   &                                   \\ \hline
MAE & - & No                  & No                   & 75                       & 1K                      & 95.56   & 70.55         & 67.35         & 38.58                             \\
MAE & - & No                  & No                   & 75                       & 3K                      & 99.62   & 71.40         & 67.67         & 39.57                             \\
MAE & - & Yes                 & Yes                  & 75                       & 10K                     & 99.87   & 71.39         & 67.70         & 40.03                             \\
MAE & - & Yes                 & Yes                  & 50                       & 10K                     & 99.70   & 80.33         & 77.34         & 23.94                             \\
MAE & - & Yes                 & Yes                  & 85                       & 10K                     & 99.61   & 66.30         & 62.10         & 47.53                             \\ 
Benchmark & 63  & - & - & - & - & 100.00 & 4.908 & 2.519 &  97.734 \\
Benchmark & 31   & - & - & - & - & 100.00 & 6.406 & 3.913 & 94.945 \\
Benchmark & 9    & - & - & - & - & 100.00 & 35.244 & 33.517 & 80.651\\
Benchmark & 7    & - & - & - & - & 100.00 & 45.714 & 43.273 & 76.177 \\
Benchmark & 3   & - & - & - & - & 100.00 &  76.613 & 74.984 & 53.457 \\ \hline
\end{tabular}
\label{tab:exp}
\end{table*}

In our experiments, we leverage the Vision Transformer (ViT) based MAE with a patch size of 14, an embedding dimension of 1280, a depth of 32 layers, 16 attention heads, and an MLP ratio of 4. The batch size was set to 128. We trained the model for a range of epochs, from 1 to 10,000, as demonstrated in Table~\ref{tab:exp}. Extended training consistently improved parroting detection rates for both seen (train) and modified samples. However, a trade-off became apparent as the training duration increased and the threshold was set based solely on training loss, leading to a model prone to flagging most samples, including the unseen ones (\(D_{\text{nov}}\)), as parroted.

\subsection{Benchmark}

We compare our parroting detection algorithm against the graph hashing approaches which are commonly used for evaluating the novelty of generated CAD models \cite{willis2021engineering,jayaraman2023solidgen,xu2023hierarchical}.  The sketch is converted into a graph with edges representing curves and nodes representing curve end points and the centers of circles and arcs.  The sketch is normalized and all point coordinates are quantized into equal sized bins.   The quantized point coordinate values are then applied to the nodes as labels while edges are labelled using the curve type.   The Weisfeiler-Lehman graph hash \cite{JMLR:v12:shervashidze11a} can then be computed.  This procedure assigns different hash values to sketches with different topology. Sketches with the same topology but small changes in geometry will receive the same graph providing the point coordinate values remain the same following quantization.  Consequently the number of bins used for quantization provides control over the amount by which the geometry can change while preserving the same hash.  In Table~\ref{tab:exp} we show results for 3, 7, 9, 31 and 63 bins. 

\subsection{Discussion}

The nuanced interplay between model parameters and the detection outcomes was evident. While weight decay and data augmentation slightly decreased the detection rates for \(D_{\text{train}}\), suggesting an impact on the model's sensitivity, a lower masking percentage (\texttt{p\_mask}) significantly enhanced detection capabilities for modified samples. Nonetheless, a higher detection rate was often accompanied by a lower \(D_{\text{nov}}\) pass percentage, indicating a potential increase in false positives. An optimal balance was observed with an 85\% \texttt{p\_mask}, which, despite a minor reduction in detection rates, resulted in the highest \(D_{\text{nov}}\) pass percentage, offering a more conservative and practical approach for scenarios where it is crucial to minimize false positives. This balance is critical in settings where the novel set may contain samples similar to the training ones, as all sets were generated from a single source.

The MAE parroting detector outperforms the detection rate of the graph based hashing benchmark for the $D_{\text{mod 1}}$ and $D_{\text{mod 2}}$ datasets, even when point coordinates are quantized into a 3 by 3 gird.   Using this very small number of bins allows very large changes to the sketch point positions, casing the hash to be largely based on the sketch topology.   As the edited sketches in our CAD variations dataset will have the same topology as the originals, graph based hashing approaches can unfairly exploit this isomorphism.   Real generated results will tend to vary from the dataset with both small changes in sketch topology and geometry being present.   Consequently the ability of the MAE parroting detector to achieve superior detection performance, with a comparable $D_{\text{nov}}$ pass percentage, and no strong dependence on sketch topology shows the utility of this technique for CAD parroting detection. 

\section{Conclusion}
\label{sec:con}

We presented a new approach for detecting generative parroting using an overfitted Masked Autoencoder with a Vision Transformer architecture. Our experiments demonstrate that training duration and model parameters, particularly the percentage mask (\texttt{p\_mask}), play significant roles in the model's ability to discern between seen, modified, and novel samples. The careful calibration of the loss threshold emerges as a crucial factor in mitigating the incidence of false positives, especially in large datasets where the distinction between original and parroted content is nuanced. Through extensive training, we observed that while a longer training time generally leads to higher detection rates for training and modified samples, it also increases the likelihood of incorrectly flagging novel samples as parroted. Our results underline the importance of selecting model configurations that balance sensitivity with specificity, thereby minimizing false positives without sacrificing detection accuracy.  The approach is shown to achieve superior performance to the hash based parroting detection algorithms employed in previous works, while not explicitly exploiting the isomorphic sketch graphs created when sketches are varied parametrically. 

Several avenues for further research are apparent. Firstly, investigating the impact of alternative architectures and learning strategies on the model's detection capabilities could yield improvements in performance. Furthermore, exploring different data modalities beyond 2D CAD sketches may provide insights into the generalizability of our approach. Additionally, the development of more sophisticated thresholding techniques that adapt to the variability within and between datasets could enhance the model's discernment between parroted and genuinely novel content. Finally, as generative models continue to evolve, ongoing collaboration with legal experts will be vital in ensuring that our detection mechanisms remain aligned with the latest copyright legislation and ethical standards. This will ensure that advancements in AI content generation move forward responsibly and sustainably.

\section*{Acknowledgments}
We would like to thank Ali Mahdavi-Amiri from Simon Fraser University and Karl D.D. Willis from Autodesk Research for the initial discussions on this work, which provided valuable insights.

{
    \small
    \bibliographystyle{ieeenat_fullname}
    \bibliography{main}
}


\end{document}